\documentclass[conference]{IEEEtran}
\IEEEoverridecommandlockouts
\usepackage{amsmath,amssymb,amsfonts}
\usepackage{algorithmic}
\usepackage{subcaption}
\usepackage{graphicx}
\usepackage{textcomp}
\usepackage{xcolor}
\usepackage{biblatex}
\usepackage{svg}
\usepackage{fancyhdr}
\bibliography{references}

\def\BibTeX{{\rm B\kern-.05em{\sc i\kern-.025em b}\kern-.08em
    T\kern-.1667em\lower.7ex\hbox{E}\kern-.125emX}}
\begin{document}

\title{Development of a Dual-Input Neural Model for Detecting AI-Generated Imagery}

\author{
    \IEEEauthorblockN{\textbf{Jonathan Gallagher}}
    \IEEEauthorblockA{
        \textit{Computational Mathematics} \\
        \textit{University of Waterloo} \\
        Email: jonathan.gallagher@uwaterloo.ca
    }
    \and
    \IEEEauthorblockN{\textbf{William Pugsley}}
    \IEEEauthorblockA{
        \textit{Computational Mathematics} \\
        \textit{University of Waterloo} \\
    }
}

\maketitle

\begin{abstract}

Over the past years, images generated by artificial intelligence have become more prevalent and more realistic. Their advent raises ethical questions relating to misinformation, artistic expression, and identity theft, among others. The crux of many of these moral questions is the difficulty in distinguishing between real and fake images. It is important to develop tools that are able to detect AI-generated images, especially when these images are too realistic-looking for the human eye to identify as fake. This paper proposes a dual-branch neural network architecture that takes both images and their Fourier frequency decomposition as inputs. We use standard CNN-based methods for both branches as described in Stuchi et al. \cite{stuchi2020frequency}, followed by fully-connected layers. Our proposed model achieves an accuracy of 94\% on the CIFAKE dataset, which significantly outperforms classic ML methods and CNNs, achieving performance comparable to some state-of-the-art architectures, such as ResNet.

\end{abstract}

\section{Introduction}

With the advent of high-quality, publicly available artificial-intelligence image generation tools such as DALL·E and Midjourney, billions of artificial images have been created and shared online. Such fake images have only become more realistic and consequently more difficult to identify as their popularity has grown. The problem is only further complicated by the fact that algorithms used in generating these images are themselves constantly and rapidly evolving. Despite efforts to reliably identify artificially generated imagery, it is often possible to circumvent technical solutions to this problem. For example, AI tools intended for identifying these images may fail to generalize when faced with images generated using different models than those used in the creation of the training set \cite{zhu2023genimage}. Researchers at Mozilla found that most current disclosure methods, such as labels and watermarks denoting an image as AI-generated, “may not prevent or effectively address harm once it has occurred” \cite{mozilla2024}. In light of this, it is important to develop new techniques that assist in making the distinction between real and fake images. Neural networks are an ideal candidate for this task because they can be trained to classify images using information that is beyond the scope of human sensory perception. Since image generating AI systems are designed to create outputs as realistic as possible, it is this information that we may have to rely on.

\begin{figure}[htbp]
\centering
\includegraphics[width=1\columnwidth]{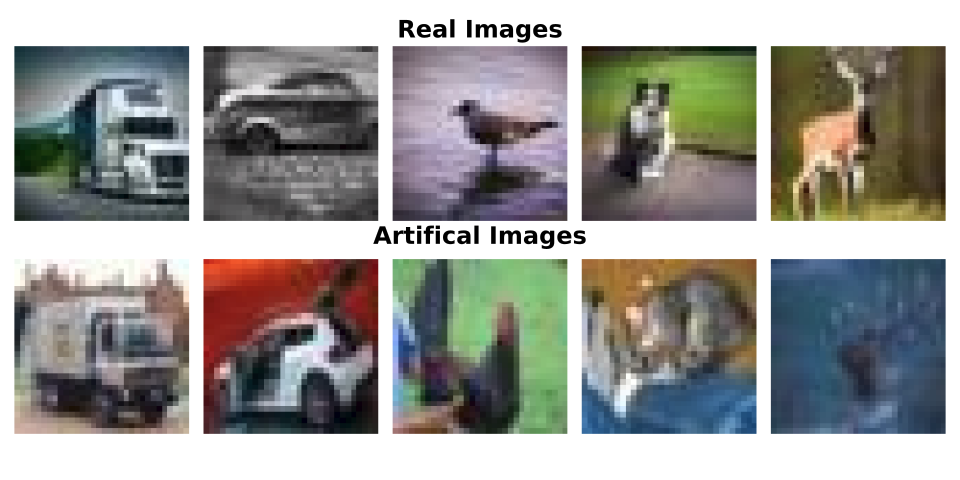}  
\caption{Real and artificially-generated images from the CIFAKE dataset.}
\label{fig:real_fake_examples}  
\end{figure}

Within the existing literature, a wide array of neural architectures have been used to detect AI-generated images \cite{he2015deep,liu2021swin,touvron2021training}. These models range from simple convolutional neural networks (CNN) to well-known image classification models that have been specifically fine-tuned for the task, such as ResNet. Results have been mixed, with even the best models failing to detect images generated using a different method than their training data \cite{zhu2023genimage}. For example, ResNet-50 is a popular CNN-based architecture used in computer vision; when applied to the task of detecting AI-generated images, it is capable of achieving an accuracy of $99\%$ when tested on images created using the same generator as the training samples \cite{zhu2023genimage}. However, it only has an accuracy of $58.6\%$ when tested on a dataset including fake images from seven other generators.

\begin{figure}
    \centering
    \begin{subfigure}{0.48\textwidth}
        \includegraphics[width=\textwidth]{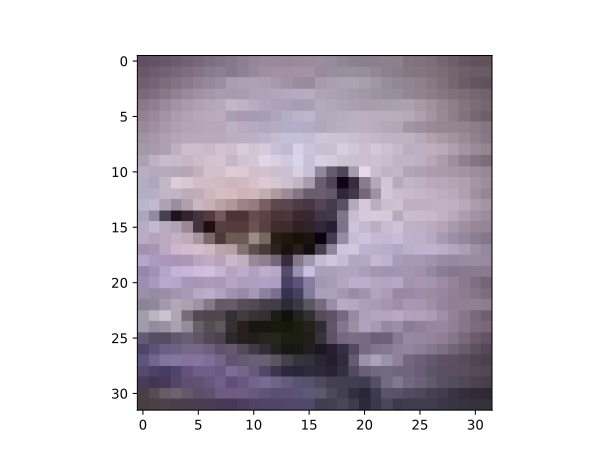}
        \caption{A cool bird.}
        \label{fig:bird_img}
    \end{subfigure}
    \hfill
    \begin{subfigure}{0.48\textwidth}
        \includegraphics[width=\textwidth]{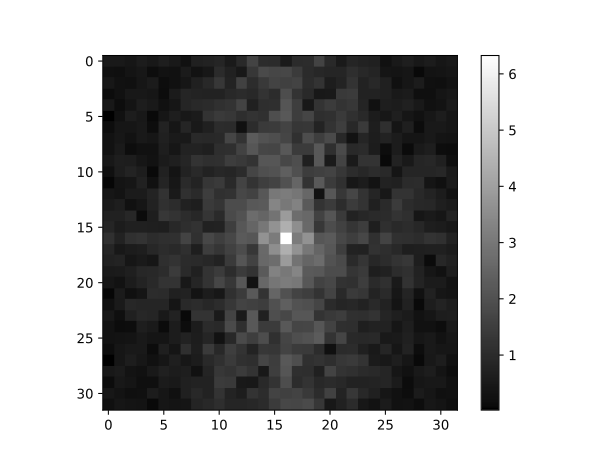}
        \caption{The $\log$-magnitude of the DFT, centered around the largest magnitude.}
        \label{fig:bird_freq}
    \end{subfigure}            
    \caption{An example image of a bird from CIFAR-10 and its corresponding DFT. For demonstration purposes, we take the DFT of the gray-scaled image}
    \label{fig:bird_ex}
\end{figure}

In contrast to existing methods, which primarily make use of either the raw pixel data or frequency spectrum data from a discrete Fourier transform (DFT) as input to the network, we propose the implementation of a dual-input architecture, integrating both the frequency content and the raw pixel data to provide a unified representation of the underlying image structure. The inclusion of frequency data can reveal features distinct from the raw data, despite both originating from the same source \cite{stuchi2020frequency}. This is due to the fact that frequency analysis addresses patterns related to the rates of change and periodicity within the data which might otherwise not be evident in the spatial domain representation. In particular, images generated using GANs show patterns that are visible to the human eye and which are generally not present in normal images \cite{zhang2019detecting,wang2020cnngenerated,gragnaniello2021gan}. Fig.~\ref{fig:bird_ex} displays a bird from the CIFAR-10 dataset and its DFT; our model will train using both these images as inputs.

\section{Model}

\begin{figure*}[h]
    \centering
    \includegraphics[width=\textwidth]{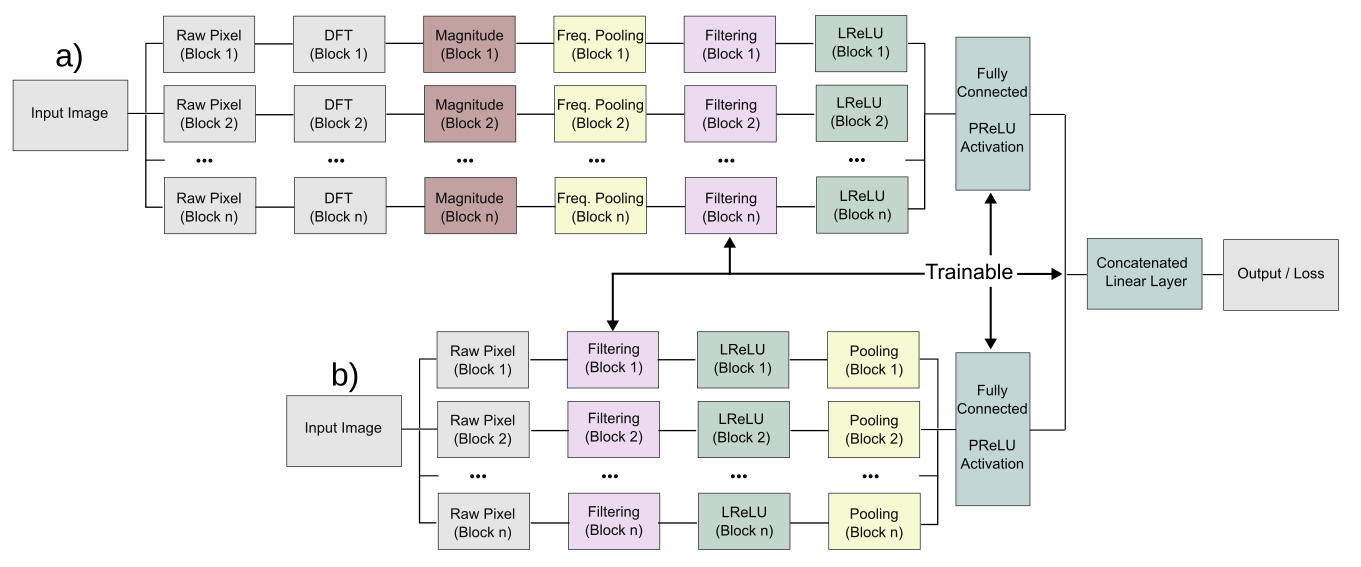} 
    \caption{Schematic of the branched network architecture depicting: a) the frequency branch and b) the image branch, where layers which serve the same purpose are color coded.}
    \label{fig:architecture}
\end{figure*}

The proposed neural network model consists of two distinct branches; the first taking a colour image as input, and the second taking the discrete Fourier transform of the same image as input. The layers in each branch do not interact with nor affect parameter values held in each others layers. Branch outputs are concatenated before being passed through further linear layers. An overview of the architecture is pictured in Fig. \ref{fig:architecture}. 

As the DFT is a transformation of the original input, choosing to provide it as additional input does not necessarily provide new information to the model. However, the DFT does in fact present the information in a manner that may allow features that would otherwise be less evident to be learned more easily. In this sense, the proposed workflow can be likened to data augmentation.

\subsection{Frequency Domain Branch}

\begin{figure}[htbp]
\centering
\includegraphics[width=1\columnwidth]{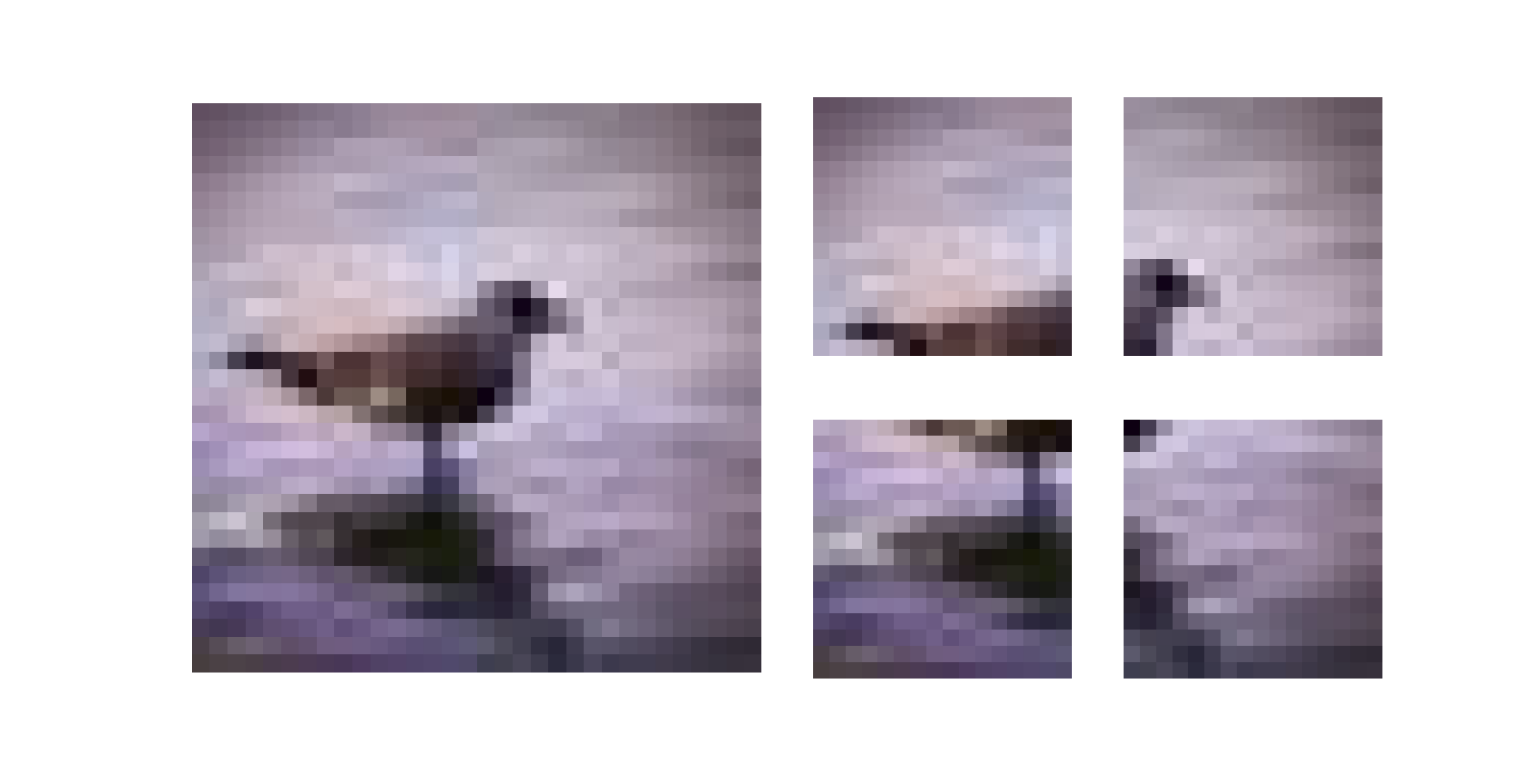}  
\caption{A $32\times32$ image of a bird split into four $16\times16$ sub-blocks.}
\label{fig:block_decomp}  
\end{figure}

We design the frequency branch following the methodology outlined by Stuchi et al. \cite{stuchi2020frequency}. The first step in the workflow involves the segmentation of image into quadrants, which will be addressed as ``blocks'' throughout the remainder of the text. If necessary, these blocks can be further segmented into sub-blocks for an even more granular approach. The purpose of this segmentation is to encourage the model to learn both local and global features within the image. For example, a $32\times32$ image can be split into four $16\times16$ blocks, which can each be subsequently split into four $8\times8$ sub-blocks, with subsequent operations applied to each block independently. With the previous example, we would have 21 sets of pixels to operate on (the entire image, its four $16\times16$ blocks, and the 16 $8\times8$ sub-blocks). Fig.~\ref{fig:block_decomp} provides a visual representation of the first stage of this process.

Prior to sectioning each image into quadrants, its content is parsed into one of three colour channels (RGB) such that the two-dimensional DFT can be applied to each channel separately. For a channel with size $A\times B$ and pixel values $f_{lj}$, the corresponding entries of the DFT are
\begin{multline}\label{eq:DFT}
    F_{hk} = \sum_{l=0}^{A-1} \sum_{j=0}^{B-1} f_{lj} e^{-2\pi i (hl/A + kj/B)},\ \text{where}\\
    h=0,1,\ldots,A-1\ \text{and}\ k=0,1,\ldots,B-1.
\end{multline}
Since the outputs of DFTs are complex-valued, the outputs of the operation are evaluated in terms of their complex magnitude, $M_{hk}$, for each frequency value to get
\begin{equation}\label{eq:magnitude}
    M_{hk}=\sqrt{(\text{Re}(F_{hk}))^2 + (\text{Im}(F_{hk}))^2},
\end{equation}
where $\text{Re}$ and $\text{Im}$ denote the real and imaginary parts of a complex number, respectively. This sacrifices some information but is necessary in order for the network to learn using back-propagation. After determining the magnitude we further take the natural logarithm of $M_{hk}$, as raw outputs may be separated by several orders of magnitude. Next, we include a max pooling layer, reducing the dimensionality of the problem space and speeding up the training process. Continuing with the previous example, our 21 blocks have been reduced to one $16\times16$ block which was originally the entire image, four $8\times8$ blocks, and 16 $4\times4$ blocks.

\begin{figure}[b]
    \centering
    \includegraphics[width=1\columnwidth]{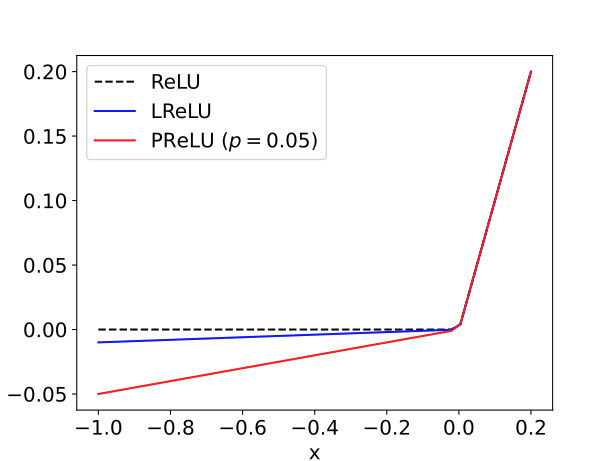}
    \caption{ReLU, LReLU, and PReLU with paramater $p=0.05$ activation functions.}
    \label{fig:act_funcs}
\end{figure}

After processing and transforming the inputs, the data is passed through a series of convolution/filtering layers. One convolution filter is used for each block size. Continuing with our example, a filter's parameters are trained on the $16\times16$ block, another on all four $8\times8$ blocks, and a third on the remaining $4\times4$ blocks. Having taken the magnitude of the frequency values, this step is no different from the convolution layers in any other CNN. After which, data is passed through a rectified linear unit (LReLU) activation function \cite{dubey2022activation}. LReLU is similar to standard ReLU, except for negative inputs where the function is linear with slope $0.01$. For input current $x$, the activation is
\begin{equation}\label{eq:LReLU}
\text{LReLU}(x)=
    \begin{cases}
        x & \text{if } x \ge 0\\
        0.01\cdot x & \text{if } x < 0
    \end{cases}.
\end{equation}
The function's kink for non-zero values helps prevent dead neurons and overall improves model training dynamics in deeper networks.

Finally, all blocks and sub-blocks are flattened and concatenated before passing the results through a series of fully-connected layers. Within these layers we opt to use the parametric rectified linear unit (PReLU) activation functions \cite{he2015delving}. PReLU is an adaptation of leaky ReLU, where the coefficient of the negative portion of the function is a learnable parameter. This change allows the activation function to adapt dynamically to the data during training. For an input current $x$, the PReLU activation is 
\begin{equation}\label{eq:PReLU}
\text{PReLU}(x)=
    \begin{cases}
        x & \text{if } x \ge 0\\
        p\cdot x & \text{if } x < 0
    \end{cases},
\end{equation}
where $p$ is a learnable parameter.

\subsection{Spatial Domain Branch}

The spatial domain branch inherits a similar architecture from the frequency branch. In a similar manner, the image is parsed into its RGB channels and divided into sub-blocks before being fed into the learnable layers. We use the same operations as described in the frequency layer, with the only exception being the absence of the DFT. In contrast to the frequency layer, we have no need to take the magnitude of the pixel values as they are not complex valued. After splitting the image into its blocks and sub-blocks, we pass these through our filtering layers followed by LReLU activation functions. The convolutions are trained on blocks which are the same size as those present in the frequency layer, however, their parameters are entirely distinct from the filters in the other branch. After filtering, max-pooling is applied to down-sample the feature map and reduce computational complexity. Similarly, the spatial branch concludes with fully-connected layers and further PReLU activation functions.

\subsection{Merged Layers}
At their intersection, the outputs from both the frequency and image branches are concatenated into a single vector, which is then passed through further fully connected layers, again using PReLU activation. The final output layer consists of a single neuron, representing the network's estimate for the probability that the input image is AI-generated. As such, the logistic activation function is applied to the output such that it is constrained to the range $[0, 1]$.

The network is architected (padding, stride,filter size, etc.) such that the outputs of both the spatial and frequency branches have identical dimensionality. This way, each branch has an equal contribution to the merged fully connected. If it is desired that one branch is weighed more heavily in the model's analysis its output can be made bigger, and vice versa.

Fig.~\ref{fig:architecture} is a diagram of the complete model architecture. The fully connected layers, filtering blocks and PReLU are highlighted as those with learnable parameters.

\begin{figure}[htb]
\centering
    \begin{subfigure}{0.48\textwidth}
        \includegraphics[width=\textwidth]{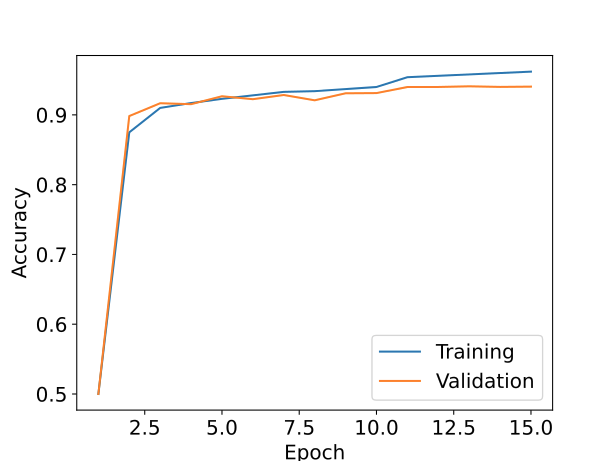}
        \caption{Accuracy of the model on the training/test set during the training loop.}
        \label{fig:training_loss}
    \end{subfigure}
    \hfill
    \begin{subfigure}{0.48\textwidth}
        \includegraphics[width=\textwidth]{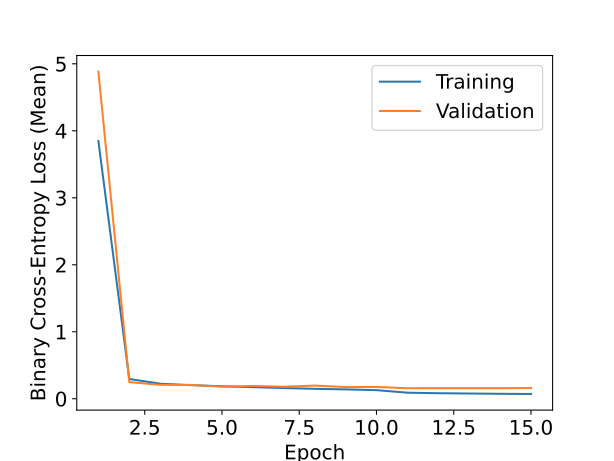}
        
        \caption{Binary cross-entropy loss of the model for the training/test set during the training loop.}
        \label{fig:training_acc}
    \end{subfigure}            
    \caption{Performance metrics of the model during training.}
    \label{fig:training_plots}
\end{figure}

\section{Experiment}

The model's performance was evaluated by training and testing using the CIFAKE dataset, comparing its performance to the performance of other state-of-the-art neural network architectures prevalent in the literature. The CIFAKE dataset was intentionally selected because of its manageable size given constraints on time and computational resources. 

The following paragraph is a summary of the relevant properties of the CIFAKE dataset, as found in the paper in which it was originally presented \cite{bird2023cifake}. CIFAKE is composed of both the classic CIFAR-10 dataset, made up of real images, and an equal number of artificial images generated using SDM, a latent diffusion model. Each image is labelled as either real (class 0) or fake (class 1). The dataset is composed of 100,000 training samples and 20,000 test samples, each composed of equal parts authentic and generated images. To provide certainty that the model is indeed learning to distinguish between real and generated images, the fake images are generated to intentionally resemble the ten classes in the original CIFAR-10 dataset. Otherwise, one could argue that the model is learning to classify images as being an instance of one of the ten classes from CIFAR-10 or an instance of whatever classes may be present in the artificial dataset.

The models which were used as benchmarks are: a classic support vector machine (SVM), a CNN, ResNet-50, VGGNet, and DenseNet \cite{wang2024harnessing}. For the last three models, Wang et al. used weights which were pre-trained on the ImageNet dataset; these weights were then fine-tuned using the CIFAKE images.

\subsection{Training}\label{sec:train}

Since the ResNet, VGGNet, and DenseNet networks are trained on ImageNet data, it is important that the images they take as inputs are processed in the same way. While we do not use any pre-trained weights, we will follow the same data pre-processing/augmentation procedures as in Wang et al. \cite{wang2024harnessing}. Our data should be as similar to that used by our benchmarks as possible so that any difference in performance can be attributed to the architecture. We horizontally flip each image with 0.5 probability and normalize over each colour channel using preset means $(0.485,\ 0.456,\ 0.406)$ and standard deviations $(0.229,\ 0.224,\ 0.225)$.

Given the relatively low resolution of the CIFAKE images ($32\times32$ pixels) we opted to split the image into one set of four $16\times16$ sub-blocks. Experimentation demonstrated that further segmentation resulted in noise in the dataset and diminished performance. Testing further divisions into $8\times8$ and $4\times4$ blocks reduced the model's accuracy by more than $0.20$, all other factors remaining equal.

The model was trained on the CIFAKE dataset for a total of $15$ epochs using the Adam optimizer and batch sizes of $32$ images. Our learning rate is scheduled such that it is held constant at $10^{-4}$ for the first ten epochs, and reduced to $10^{-5}$ for the last five. As the network is intended to perform binary classification it is suitable to implement binary cross-entropy as our loss function. 

A variety of regularization techniques were implemented in order to improve the solution quality and prevent overfitting:
\begin{itemize}
    \item \textbf{Dropout}: Dropout was applied to each unit in the linear layers with probability $0.5$. Dropout was not applied to the convolutional layers.
    \item \textbf{Weight Initialization}: Uniform Glorot initialization was implemented for all the linear layers' weights and convolution layers' filters. All biases were initialized to zero.
    \item \textbf{Data Augmentation}: As described previously, a random horizontal flip was applied to the images before being normalized.
    \item \textbf{Scheduled Learning Rate}: Learning rate was scheduled such that it decreases by an order of magnitude every ten epochs to help the parameters stabilize around an equilibrium.
\end{itemize}

\section{Results}

After training, the model achieved a test accuracy of $0.94$. Fig.~\ref{fig:training_plots} shows the performance of the model throughout the training process. It is observed that the training and test loss both decrease steadily during the first ten epochs, after which the learning rate is reduced by an order of magnitude, as detailed in section~\ref{sec:train}. For remaining five epochs, the test loss stays constant while the training loss slightly decreases. This is indicative of the model converging upon a stable solution and not overfitting. Fig.~\ref{fig:training_acc} corroborates this conclusion; the training and test accuracies steadily improve before the test accuracy settles at an equilibrium solution while the training accuracy grows slightly. 

\begin{table}[t]
    \centering
    \caption{Results for SVM, CNN, ResNet, VGGNet, and DenseNet models from Wang et al. \cite{wang2024harnessing}.}
    \label{tab:comparison}
    \begin{tabular}{c||c c c c}
        Model & Precision & Recall & F1-Score & Accuracy \\
        \hline
        SVM & 0.8020 & 0.8222 & 0.8120 & 0.8143 \\
        CNN & 0.8734 & 0.8574 & 0.8653 & 0.8640 \\
        ResNet & 0.9917 & 0.9066 & 0.9472 & 0.9495  \\
        VGGNet & 0.9657 & 0.9547 & 0.9602 & 0.9600  \\
        DenseNet & 0.9769 & 0.9779 & 0.9774 & 0.9774 \\
        \hline
        \textbf{Our Model} & 0.9444 & 0.9321 & 0.9570 & 0.9451
    \end{tabular}
\end{table}

A detailed account of the model's performance in comparison to the benchmarks can be found in TABLE~\ref{tab:comparison}. We find that our model outperforms both the SVM and CNN models. The original CIFAKE paper proposed another CNN-based model that achieved a test accuracy of just under 0.93, which our model also outperforms \cite{bird2023cifake}. However, our model underperforms in all metrics when compared to VGGNet and DenseNet. Our model is most similar to ResNet in performance; the accuracies are within $0.01$ of each other and the F1-scores are even closer. To understand this result we need to recognize that ResNet's precision is higher, but its recall is lower. Since the F1-score of a classifier is the harmonic mean of its precision and recall, it is sensible that our values average to similar results despite their discrepancies.

ResNet has high precision. Indeed, it is higher than any other model's we consider and higher than any value for recall. In the context of AI image detection, this means that when ResNet classifies a picture as being AI-generated, it is very likely to be correct. However, its relatively low recall presents an issue. There will be proportionally more AI-generated images that go undetected when we use ResNet over our model. Determining which of these two models is more suitable will depend on the specific application and problem context. For example, in some contexts it may be preferable to detect many fake images even at the expense of increasing the rate of false positives. The choice will require careful consideration as to whether or not the harm from incorrectly classifying a real image as fake is less than the harm from incorrectly classifying a fake image as real. This is a very situational question and will depend for what purpose the user wishes to distinguish between AI-generated and real images.

\section{Conclusion}
This study introduced a novel dual-branch neural network architecture, incorporating both image and discrete Fourier frequency decompositions to detect AI-generated images with high reliability. In comparison to established benchmarks, our model achieved an accuracy of $0.94$, outperforming conventional machine learning and CNN-based approaches, and demonstrating performance which is comparable to that of state of the art models such as ResNet. 

In its current state, the proposed architecture is not yet comparable to VGGNet and DenseNet, falling short in all metrics. Despite this, a more systematic approach to selecting and tuning hyper-parameters could potentially result in further improve performance, provided a longer time horizon.

Future work should investigate the application of the model in a broader array of contexts, such as how it performs on images which were generated using methods that were not included in the training set. Doing so will allow for an understanding of how generalizable our model is and where necessary improvements can be made. Furthermore, the model would benefit from a more sophisticated dataset, ideally one which more closely mimics the evolving capabilities of state of the art image generation techniques. With CIFAKE, we are limited to low resolution images of ten kinds of objects. Such a dataset will ensure that we are assessing the models utility in real world scenarios, where generative techniques are ever-evolving. 

Going beyond the dataset, we can make changes to the architecture as well. Our use of standard CNN tools is not the only approach. We can replace either of the branches with residual neural networks, transformers, or many others. Even had our model surpassed previous results, it would likely not remain in the top spot for long. The use of Fourier decompositions is one more tool for researchers and developers to consider. As the landscape of AI-generated images continues to change, so to must the architectures and tools we use to respond to it.

\printbibliography 

@misc{he2015deep,
      title={Deep Residual Learning for Image Recognition}, 
      author={Kaiming He and Xiangyu Zhang and Shaoqing Ren and Jian Sun},
      year={2015},
      eprint={1512.03385},
      archivePrefix={arXiv},
}

@misc{touvron2021training,
      title={Training Data-Efficient Image Transformers \& Distillation through attention}, 
      author={Hugo Touvron and Matthieu Cord and Matthijs Douze and Francisco Massa and Alexandre Sablayrolles and Hervé Jégou},
      year={2021},
      eprint={2012.12877},
      archivePrefix={arXiv},
}

@misc{liu2021swin,
      title={Swin Transformer: Hierarchical Vision Transformer using Shifted Windows}, 
      author={Ze Liu and Yutong Lin and Yue Cao and Han Hu and Yixuan Wei and Zheng Zhang and Stephen Lin and Baining Guo},
      year={2021},
      eprint={2103.14030},
      archivePrefix={arXiv},
}

@misc{wang2020cnngenerated,
      title={CNN-Generated Images Are Surprisingly Easy to Spot... For Now}, 
      author={Sheng-Yu Wang and Oliver Wang and Richard Zhang and Andrew Owens and Alexei A. Efros},
      year={2020},
      eprint={1912.11035},
      archivePrefix={arXiv},
}

@misc{zhang2019detecting,
      title={Detecting and Simulating Artifacts in GAN Fake Images}, 
      author={Xu Zhang and Svebor Karaman and Shih-Fu Chang},
      year={2019},
      eprint={1907.06515},
      archivePrefix={arXiv},
}

@misc{stuchi2020frequency,
      title={Frequency Learning for Image Classification}, 
      author={José Augusto Stuchi and Levy Boccato and Romis Attux},
      year={2020},
      eprint={2006.15476},
      archivePrefix={arXiv},
}

@misc{zhu2023genimage,
      title={GenImage: A Million-Scale Benchmark for Detecting AI-Generated Image}, 
      author={Mingjian Zhu and Hanting Chen and Qiangyu Yan and Xudong Huang and Guanyu Lin and Wei Li and Zhijun Tu and Hailin Hu and Jie Hu and Yunhe Wang},
      year={2023},
      eprint={2306.08571},
      archivePrefix={arXiv},
}

@online{mozilla2024,
    title={In Transparency We Trust?},
    author={Ramak Molavi Vasse'i and Gabriel Udoh},
    year={2024},
    month={02},
    day={26},
    url={https://foundation.mozilla.org/en/research/library/in-transparency-we-trust/research-report/#key-findings},
    publisher={Mozilla},
}

@misc{wang2024harnessing,
      title={Harnessing Machine Learning for Discerning AI-Generated Synthetic Images}, 
      author={Yuyang Wang and Yizhi Hao and Amando Xu Cong},
      year={2024},
      eprint={2401.07358},
      archivePrefix={arXiv},
      primaryClass={cs.CV}
}

@misc{bird2023cifake,
      title={CIFAKE: Image Classification and Explainable Identification of AI-Generated Synthetic Images}, 
      author={Jordan J. Bird and Ahmad Lotfi},
      year={2023},
      eprint={2303.14126},
      archivePrefix={arXiv}
}

@misc{he2015delving,
      title={Delving Deep into Rectifiers: Surpassing Human-Level Performance on ImageNet Classification}, 
      author={Kaiming He and Xiangyu Zhang and Shaoqing Ren and Jian Sun},
      year={2015},
      eprint={1502.01852},
      archivePrefix={arXiv},
      primaryClass={cs.CV}
}

@article{dubey2022activation,
    title = {Activation functions in deep learning: A comprehensive survey and benchmark},
    journal = {Neurocomputing},
    volume = {503},
    pages = {92-108},
    year = {2022},
    issn = {0925-2312},
    doi = {https://doi.org/10.1016/j.neucom.2022.06.111},
    url = {https://www.sciencedirect.com/science/article/pii/S0925231222008426},
    author = {Shiv Ram Dubey and Satish Kumar Singh and Bidyut Baran Chaudhuri}
}

@misc{gragnaniello2021gan,
    author = {Gragnaniello, Diego and Cozzolino, Davide and Marra, Francesco and Poggi, Giovanni and Verdoliva, Luisa},
    year={2021},
    eprint={2104.02617},
    archivePrefix={arXiv},
    title = {Are GAN Generated Images Easy to Detect? A Critical Analysis of the State-of-the-Art}
}
\end{document}